\UseRawInputEncoding
\UseRawInputEncoding
\documentclass[final]{article}

\PassOptionsToPackage{numbers,sort&compress}{natbib}
\usepackage{neurips_2025}
\usepackage[utf8]{inputenc} 
\usepackage[T1]{fontenc}    
\usepackage{hyperref}       
\usepackage{url}            
\usepackage{booktabs}       
\usepackage{colortbl}
\usepackage{amsfonts}       
\usepackage{nicefrac}       
\usepackage{microtype}      
\usepackage{xcolor}         %
\usepackage{graphicx}       
\usepackage{algorithm}
\usepackage{algpseudocode}
\usepackage{tikz}
\usepackage{amsmath}
\usepackage{amssymb}
\usepackage{fontawesome5}
\usepackage{xspace}
\usetikzlibrary{calc}

\definecolor{algborder}{HTML}{C8CED6}
\definecolor{algheader}{HTML}{F0F3F6}
\definecolor{algheadertext}{HTML}{2B3440}
\definecolor{fn}{HTML}{1565A8}
\definecolor{accent}{HTML}{C26522}
\definecolor{footertext}{HTML}{57606A}
\definecolor{rowalt}{HTML}{F8F9FB}
\definecolor{bestcell}{HTML}{EDF4FB}

\providecommand{\best}[1]{\cellcolor{bestcell}\textcolor{fn}{\textbf{#1}}}

\definecolor{algheader}{HTML}{F0F3F6}
\definecolor{algheadertext}{HTML}{2B3440}
\definecolor{initcolor}{HTML}{EFF6EE}
\definecolor{loopcolor}{HTML}{FDF6EC}
\definecolor{evalcolor}{HTML}{EDF4FB}
\definecolor{returncolor}{HTML}{F3EFF6}
\definecolor{accent}{HTML}{C26522}

\algrenewcommand\algorithmicfor{\textcolor{kw}{\textbf{for}}}
\algrenewcommand\algorithmicif{\textcolor{kw}{\textbf{if}}}
\algrenewcommand\algorithmicelse{\textcolor{kw}{\textbf{else}}}
\algrenewcommand\algorithmicend{\textcolor{kw}{\textbf{end}}}
\algrenewcommand\algorithmicdo{\textcolor{kw}{\textbf{do}}}
\algrenewcommand\algorithmicreturn{\textcolor{kw}{\textbf{return}}}
\algrenewcommand\algorithmicthen{}
\algrenewcommand\algorithmiccomment[1]{\hfill\textcolor{cmt}{\small\textit{#1}}}

\providecommand{\phase}[2]{}%
\renewcommand{\phase}[2]{%
  \vspace{4pt}%
  \noindent\colorbox{#1}{\parbox{\dimexpr\linewidth-2\fboxsep}{%
    \vspace{1.5pt}%
    \hspace{3pt}{\footnotesize\textcolor{footertext}{\textsf{#2}}}%
    \vspace{1.5pt}%
  }}%
  \vspace{2pt}%
}
\newcommand{\github}{\raisebox{-0.5pt}{\includegraphics[height=1.05em]{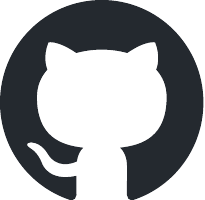}}\xspace}
\definecolor{algheader}{HTML}{F0F3F6}
\definecolor{algheadertext}{HTML}{2B3440}
\definecolor{initcolor}{HTML}{EFF6EE}
\definecolor{loopcolor}{HTML}{FDF6EC}
\definecolor{evalcolor}{HTML}{EDF4FB}
\definecolor{returncolor}{HTML}{F3EFF6}
\definecolor{accent}{HTML}{C26522}

\algrenewcommand\algorithmicfor{\textcolor{kw}{\textbf{for}}}
\algrenewcommand\algorithmicif{\textcolor{kw}{\textbf{if}}}
\algrenewcommand\algorithmicelse{\textcolor{kw}{\textbf{else}}}
\algrenewcommand\algorithmicend{\textcolor{kw}{\textbf{end}}}
\algrenewcommand\algorithmicdo{\textcolor{kw}{\textbf{do}}}
\algrenewcommand\algorithmicreturn{\textcolor{kw}{\textbf{return}}}
\algrenewcommand\algorithmicthen{}
\algrenewcommand\algorithmiccomment[1]{\hfill\textcolor{cmt}{\small\textit{#1}}}

\providecommand{\phase}[2]{}%
\renewcommand{\phase}[2]{%
  \vspace{4pt}%
  \noindent\colorbox{#1}{\parbox{\dimexpr\linewidth-2\fboxsep}{%
    \vspace{1.5pt}%
    \hspace{3pt}{\footnotesize\textcolor{footertext}{\textsf{#2}}}%
    \vspace{1.5pt}%
  }}%
  \vspace{2pt}%
}

\definecolor{algborder}{HTML}{CBD5E1}
\definecolor{headerbar}{HTML}{334155}
\definecolor{headertext}{HTML}{FFFFFF}
\definecolor{kw}{HTML}{6D28D9}
\definecolor{fn}{HTML}{1D4ED8}
\definecolor{cmt}{HTML}{94A3B8}
\definecolor{phase}{HTML}{64748B}
\definecolor{accentbar}{HTML}{F59E0B}
\definecolor{footertext}{HTML}{64748B}

\algrenewcommand\algorithmicfor{\textcolor{kw}{\textbf{for}}}
\algrenewcommand\algorithmicif{\textcolor{kw}{\textbf{if}}}
\algrenewcommand\algorithmicelse{\textcolor{kw}{\textbf{else}}}
\algrenewcommand\algorithmicend{\textcolor{kw}{\textbf{end}}}
\algrenewcommand\algorithmicdo{\textcolor{kw}{\textbf{do}}}
\algrenewcommand\algorithmicreturn{\textcolor{kw}{\textbf{return}}}
\algrenewcommand\algorithmicthen{}
\algrenewcommand\algorithmiccomment[1]{\hfill\textcolor{cmt}{\(\triangleright\) \small\textit{#1}}}

\newcommand{\fn}[1]{\textcolor{fn}{\textsc{#1}}}

\providecommand{\phase}[1]{%
  \vspace{6pt}%
  \noindent
  \textcolor{accentbar}{\rule{2.5pt}{9pt}}\;\,
  {\small\sffamily\textcolor{phase}{\textbf{#1}}}%
  \vspace{3pt}%
}

\makeatletter
\renewcommand{\ALG@beginalgorithmic}{\small}
\makeatother
\usepackage{caption}
\usepackage{float}
\usepackage{placeins}
\usepackage{listings}

\definecolor{codebg}{HTML}{F5F5F5}
\definecolor{border}{HTML}{D0D0D0}
\definecolor{skillhead}{HTML}{FFF3E0}
\definecolor{pyhead}{HTML}{E3F2FD}
\definecolor{codecomment}{HTML}{6E7781}
\definecolor{codekw}{HTML}{CF222E}
\definecolor{codestr}{HTML}{0A3069}

\lstdefinestyle{base}{
  basicstyle=\ttfamily\footnotesize,
  backgroundcolor=\color{codebg},
  breaklines=true,
  columns=fullflexible,
  keepspaces=true,
  frame=single,
  rulecolor=\color{border},
  framerule=0.4pt,
  framesep=5pt,
  xleftmargin=5pt,
  xrightmargin=5pt,
  aboveskip=0pt,
}

\lstdefinestyle{md}{
  style=base,
  mathescape=false,
  literate={\$}{{\textdollar}}1,
  escapeinside={(*@}{@*)},
}

\lstdefinestyle{py}{
  style=base,
  language=Python,
  keywordstyle=\color{codekw}\bfseries,
  stringstyle=\color{codestr},
  commentstyle=\color{codecomment}\itshape,
  numbers=left,
  numberstyle=\tiny\color{codecomment},
  numbersep=10pt,
  showstringspaces=false,
}

\newcommand{\filehead}[2]{%
  \par\noindent
  \setlength{\fboxrule}{0.4pt}%
  \fcolorbox{border}{#1}{\parbox{\dimexpr\linewidth-2\fboxsep-2\fboxrule}{%
    \small\ttfamily\bfseries\strut #2%
  }}%
  \vspace{-2.4pt}%
}

\newcommand{\mdrender}[2]{%
  \filehead{skillhead}{#1}%
  \begingroup%
  \catcode`\$=12\relax
  \catcode`\_=12\relax
  \lstinputlisting[style=md]{#2}%
  \endgroup%
}

\newcommand{\pyrender}[2]{%
  \filehead{pyhead}{#1}%
  \lstinputlisting[style=py]{#2}%
}

\title{EvoSkill: Automated Skill Discovery for \\ Multi-Agent Systems}

\author{%
  Salaheddin Alzubi$^{1}$\thanks{Preprint. Work in progress.}\\[2pt]
  \textbf{Noah Provenzano}$^{2}$\qquad \textbf{Jaydon Bingham}$^{2}$\qquad \textbf{Weiyuan Chen}$^{2}$\qquad \textbf{Tu Vu}$^{2}$\\[2pt]
  $^{1}$Sentient,\ $^{2}$Virginia Tech\\
}

\renewenvironment{abstract}%
{%
  \vskip 0.075in%
  \centerline{\large\bf Abstract}%
  \vspace{0.5ex}%
  \begin{quote}\small%
}
{%
  \par\end{quote}%
  \vskip 0.5ex%
}

\begin{document}
\raggedbottom

\maketitle

\begin{center}

    \github \url{https://github.com/sentient-agi/EvoSkill}
\end{center}
\begin{abstract}
Coding agents are increasingly used as general-purpose problem solvers, but their flexibility does not by itself confer the domain expertise needed for specialized tasks. Recent work addresses this through \textit{agent skills}: reusable workflows, and code, that augment agents with domain-specific capabilities. Most skills today are hand-crafted, and existing evolutionary approaches optimize low-level artifacts (e.g. prompts \& code) that are tightly coupled to specific models and tasks. We introduce \textbf{EvoSkill}, a self-evolving framework that automatically discovers and refines agent skills through iterative failure analysis. EvoSkill analyzes execution failures, proposes new skills or edits to existing ones, and materializes them into structured, reusable skill folders. A Pareto frontier of agent programs governs selection, retaining only skills that improve held-out validation performance while the underlying model remains frozen. We evaluate EvoSkill on two benchmarks: OfficeQA, a grounded reasoning benchmark over U.S.\ Treasury data, where it improves exact-match accuracy by \textbf{7.3\%} (60.6\% $\to$ 67.9\%); and SealQA, a search-augmented QA benchmark with noisy retrieval, where it yields a \textbf{12.1\%} gain (26.6\% $\to$ 38.7\%). We also investigate the zero-shot transfer capabilties of skills evolved on one task to the other; in particular: skills evolved from SealQA transfers zero-shot to BrowseComp, improving accuracy by \textbf{5.3\%} without modification demonstrating that skill-level optimization produces transferable capabilities beyond the training task.
\end{abstract}

\section{Introduction}

Coding agents (e.g., Claude Code\cite{claude_code_overview}, OpenHands \cite{wang2025openhandsopenplatformai}, Codex\cite{openai_codex}) have emerged as a dominant paradigm for solving tasks across a wide range of domains. This trend is driven by the increasing use of code as a flexible intermediate representation, enabling coding agents to invoke complex abstractions and operate as general-purpose problem solvers. However, while this flexibility allows agents to interface with diverse tools and domains, it does not by itself confer the domain expertise required to perform specialized tasks at a consistently high level.

To bridge this gap, recent work has explored \textit{agent skills}: reusable, domain-specific capabilities that augment general-purpose coding agents with structured workflows, instructions, and supporting code. Most skills today are hand-crafted on an ad-hoc basis, requiring both domain knowledge and significant manual effort; a process that scales poorly as the number of target tasks grows. Evolutionary methods such as AlphaEvolve \cite{novikov2025alphaevolvecodingagentscientific}, and GEPA \cite{agrawal2026gepareflectivepromptevolution}, offer a promising alternative by optimizing agent artifacts: codebases, or prompts, through iterative search. However, these approaches operate at the \textit{artifact level}: the optimized prompts or code are tightly coupled to a specific model and task configuration, and do not naturally yield reusable components that transfer across settings.

In this work, we introduce \textbf{EvoSkill}, a self-evolving framework that operates at a higher level of abstraction: rather than optimizing prompts or code directly, EvoSkill iteratively \textit{discovers and refines agent skills} through failure-driven textual feedback. EvoSkill maintains a Pareto frontier of agent programs and, at each iteration, analyzes execution failures to propose new skills or refine existing ones. Proposed skills are materialized into structured, reusable skill folders comprising instructions, trigger metadata, and helper scripts, and are retained only if they improve performance on a held-out validation set. Skills accumulate across iterations, progressively expanding the agent's capabilities while the underlying model remains frozen.

We validate EvoSkill across two benchmarks. On \textbf{OfficeQA}\cite{databricks_officeqa_2025}, a grounded reasoning benchmark over U.S.\ Treasury data, EvoSkill improves Claude Code with Opus 4.5 from 60.6\% to \textbf{67.9\%} exact-match accuracy (+7.3\%) using only a small training subset. On \textbf{SealQA}\cite{pham2025sealqaraisingbarreasoning}, a search-augmented QA benchmark with noisy retrieval, EvoSkill yields a \textbf{12.1\%} improvement (26.6\% $\to$ 38.7\%). Furthermore, a skill evolved on SealQA transfers zero-shot to \textbf{BrowseComp}\cite{wei2025browsecompsimplechallengingbenchmark} with no modifications, improving accuracy by \textbf{5.3\%} providing direct evidence that skills discovered by EvoSkill generalize beyond their training task.

Our contributions are as follows:
\begin{enumerate}
    \item We propose \textbf{EvoSkill}, a framework for automatically discovering and refining reusable agent skills through iterative failure analysis, operating at the skill abstraction level rather than on low-level artifacts such as prompts or codebases.
    \item We demonstrate that EvoSkill yields substantial improvements across two distinct benchmarks: \textbf{+7.3\%} on OfficeQA \cite{databricks_officeqa_2025}(grounded document reasoning) and \textbf{+12.1\%} on SealQA \cite{pham2025sealqaraisingbarreasoning}(search-augmented QA), using only small training subsets.
    \item We show that skills evolved by EvoSkill transfer zero-shot to unseen tasks, with a skill discovered on SealQA improving BrowseComp accuracy by \textbf{5.3\%} without modification; demonstrating that skill-level optimization produces transferable capabilities.
\end{enumerate}
\section{Methodology}

The core idea behind EvoSkill is to iteratively discover and refine agent skills by applying textual feedback descent \cite{lee2025feedbackdescentopenendedtext} to examples where the current agent fails. EvoSkill assumes a coding agent harness that supports skill folders (e.g., Claude Code, Codex, OpenCode) and a model capable of utilizing such skills. The underlying model remains frozen throughout; only the skill repository and agent metadata evolve across iterations.

\subsection{Framework Overview}

EvoSkill consists of three collaborating agents:

\begin{enumerate}
    \item \textbf{Executor Agent ($A$):} Executes tasks under the governance of the current agent program. The base program initializes the Executor with no skills.
    \item \textbf{Proposer Agent ($P$):} Analyzes the Executor's output traces, predicted answers, and ground-truth answers to diagnose failures and propose high-level skill descriptions. Ground-truth answers are provided to enable root-cause diagnosis, analogous to examining labeled misclassifications during error analysis in supervised learning, and arenot propagated to the generated skills themselves. The Proposer determines whether to create a new skill or refine an existing one.
    \item \textbf{Skill-Builder Agent ($S$):} Materializes a high-level proposal from the Proposer into a concrete skill folder comprising trigger metadata, procedural instructions (\texttt{SKILL.md}), and optional helper scripts (Python or Typescript code) or reference material. The Skill-Builder is bootstrapped with a meta-skill that codifies best practices for skill authoring.
\end{enumerate}

All agents have read access to the base agent's repository; only the Skill-Builder has write permissions to the skills directory. The Proposer additionally maintains a cumulative \textbf{feedback history} $\mathcal{H}$ that logs all prior proposals, their outcomes, and score deltas. This serves two purposes: it prevents redundant proposals, and it enables the Proposer to refine what previously partially worked and avoid making the same mistakes making its context progressively richer across iterations.

\subsection{EvoSkill Loop}

\begin{figure*}[t]
  \centering
  \setlength{\abovecaptionskip}{2pt}
  \setlength{\belowcaptionskip}{0pt}
  \includegraphics[width=0.80\textwidth]{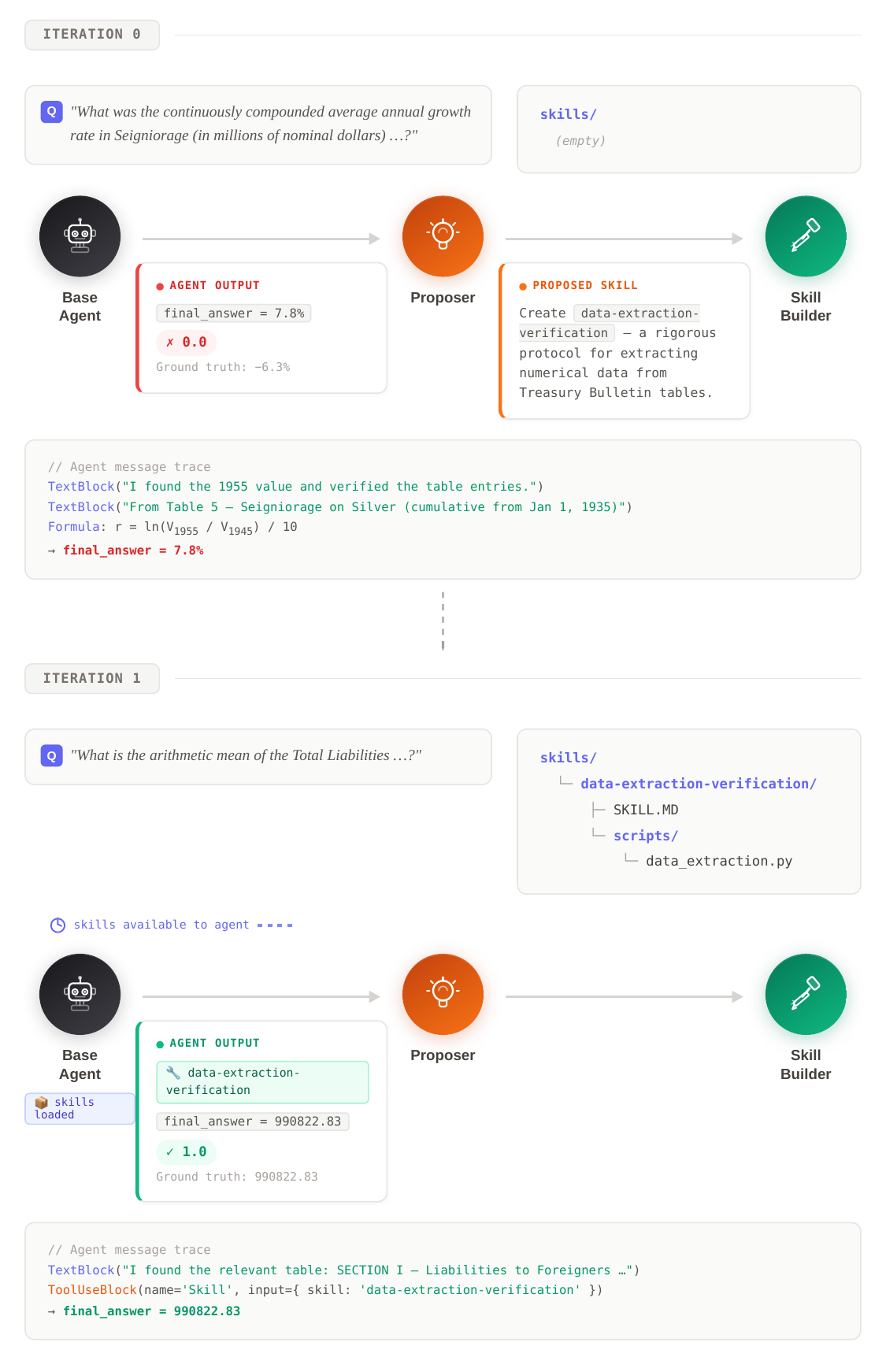}
  \caption{Overview of the EvoSkill loop.}
  \label{fig:evoskill}
\end{figure*}

\begin{algorithm}[t]
\begin{tikzpicture}
\caption{}
\label{alg:skill-loop}

\node[inner sep=0pt, outer sep=0pt] (alg) {
\begin{minipage}{\linewidth}

\noindent
\colorbox{algheader}{\parbox{\dimexpr\linewidth-2\fboxsep}{%
  \vspace{4pt}%
  \centering\color{algheadertext}%
  {\small\textsf{\textbf{Algorithm 1}}\quad
  \textcolor{accent}{EvoSkill}\;---\;Iterative Skill Induction via Textual Feedback}%
  \vspace{4pt}%
}}

\vspace{3pt}

\begin{minipage}{\dimexpr\linewidth-12pt}
\begin{algorithmic}[1]
\footnotesize

\Require
Executor agent $A$, proposer $P$,
skill-builder $S$,
evaluation set $V$, frontier capacity $k$, max iterations $T$

\vspace{1pt}
\noindent\textcolor{algborder}{\rule{\dimexpr\linewidth}{0.3pt}}
\vspace{1pt}

\phase{initcolor}{Initialization}

\State $H \gets [\,]$
    \Comment{feedback history}
\State $G \gets \{A\}$
    \Comment{frontier of top-$k$ programs}
\State $s_A \gets \fn{Eval}(A, V)$
    \Comment{baseline score}

\phase{loopcolor}{Evolution Loop}

\For{$t = 1$ \textcolor{kw}{\textbf{to}} $T$}

    \State $i \gets (t \bmod |G|)$;\; $p \gets G[i]$
        \Comment{round-robin parent selection}
    \State $F \gets \fn{CollectFailures}(p)$
        \Comment{scores $< \tau$}
    \If{$F = \emptyset$}
        \textbf{\textcolor{kw}{continue}}
    \EndIf

    \vspace{1pt}
    \State $\pi \gets P(F, H)$
        \Comment{propose new skill or edit}
    \State $\tilde{p} \gets S(p, \pi)$
        \Comment{build candidate}

    \vspace{1pt}
    \Statex\phase{evalcolor}{Frontier Update}

    \State $\tilde{s} \gets \fn{Eval}(\tilde{p}, V)$
        \Comment{held-out score}

    \If{$|G| < k$ \textcolor{kw}{\textbf{or}} $\tilde{s} > \min_{q \in G}\; \fn{Score}(q)$}
        \State $G \gets G \cup \{\tilde{p}\}$
        \Comment{Update frontier}
        \If{$|G| > k$}
            $G \gets G \setminus \{\arg\min_{q \in G}\; \fn{Score}(q)\}$
        \EndIf
    \EndIf

    \vspace{1pt}
    \State \fn{AppendFeedback}$(H, \pi, \tilde{s})$
    \Comment{Update history log}

\EndFor

\phase{returncolor}{Output}

\State \Return $\arg\max_{q \in G}\; \fn{Score}(q)$
\Comment{Return best performing program}

\end{algorithmic}
\end{minipage}

\vspace{3pt}
\noindent\textcolor{algborder}{\rule{\linewidth}{0.3pt}}
\vspace{2pt}

{\scriptsize\color{footertext}
\noindent EvoSkill maintains a frontier $G$ of the $k$ best programs.
Each iteration: a parent is selected round-robin from $G$, a proposer diagnoses its failures, a skill-builder produces a candidate,
and the candidate enters the frontier if it outscores the weakest member.}
\end{minipage}
};

\draw[algborder, rounded corners=6pt, line width=0.5pt]
  (alg.north west) rectangle (alg.south east);

\end{tikzpicture}

\end{algorithm}

EvoSkill optimizes agent programs through iterative skill mutations guided by textual feedback (Algorithm~\ref{alg:skill-loop}). A \textit{program} $p$ encapsulates the agent's system prompt and accumulated skills. The loop maintains a fixed-capacity frontier $G$ of the $k$ highest-scoring programs.

At each iteration $t$, a parent program $p$ is selected from the frontier $G$ via round-robin cycling, ensuring each frontier member is explored before any is revisited. The parent is evaluated on a training batch sampled without replacement, cycling through all examples before repeating. Responses are scored against ground-truth answers using a task-specific scoring function. Samples scoring below a fixed threshold are collected into a failure set $F$; if no failures are found, the iteration is skipped.

The Proposer $P$ receives $F$ together with the feedback history $H$ and performs structured failure analysis: reviewing execution traces, identifying capability gaps, and auditing existing skills. It then produces a textual proposal $\pi$ specifying either a new skill or an edit to an existing one.

The Skill-Builder $S$ receives the current parent program $p$ and proposal $\pi$, and materializes the proposal into a candidate program $\tilde{p}$: the parent's configuration augmented with the new or revised skill. The candidate is evaluated on the held-out validation set $V$ using the same scoring function.

The candidate enters the frontier $G$ if its score exceeds that of the weakest frontier member, displacing it; otherwise the candidate is discarded. Regardless of outcome, the proposal, its score, and the selection verdict are appended to $H$, ensuring the Proposer can reference which strategies succeeded or regressed and why. After $T$ iterations, the loop returns the highest-scoring program in $G$.

\subsection{Implementation Details}

\subsubsection{Data Setup}
\label{methodology:data_setup}

Given a supervised dataset $\mathcal{D} = \{(x_i, y_i)\}_{i=1}^{N}$ and a scoring function, we first cluster the dataset into $K$ categories using an LLM as a classifier, assigning each example to a single category. We then perform stratified partitioning into three disjoint subsets: a \textbf{training set} used for failure detection during evolution, a \textbf{validation set} used for scoring candidate programs (frontier selection), and a \textbf{held-out test set} comprising all remaining examples, which is never exposed during evolution and is used exclusively for final evaluation. Split ratios are configurable, with defaults ensuring every category is represented in both the training and validation partitions regardless of category size. Training data are organized as category-keyed pools to support the category-aware sampling used during evolution. Full details of the splitting procedure are provided in Appendix~\ref{appendix:data_setup}.

\subsubsection{Environment Setup}

EvoSkill operates within a git repository where the codebase is fixed. Each agent program is represented as a branch that diverges from its parent only in its skill folders and metadata (system prompt, lineage information, validation score). This design ensures that performance differences between programs are attributable solely to their evolved skills, while keeping program branches lightweight. Full details of the repository configuration are provided in Appendix~\ref{appendix:env_setup}.  

\section{Experiments}

We evaluate EvoSkill along three axes: (1) whether iterative skill evolution improves agent performance on challenging benchmarks, (2) what properties of the training setup influence skill quality, and (3) whether evolved skills transfer zero-shot to unseen tasks. We additionally present qualitative examples of discovered skills to illustrate the nature of the capabilities EvoSkill produces.

\subsection{OfficeQA}

\subsubsection{Benchmark}

OfficeQA is a grounded reasoning benchmark built from U.S.\ Treasury Bulletins---a corpus of approximately 89,000 pages spanning five decades of monthly and quarterly publications. Each bulletin is 100--200 pages of prose, complex tables, charts, and figures describing Treasury operations. The benchmark consists of 246 questions organized into easy and hard difficulty levels. Questions require locating and synthesizing information across an average of two bulletin documents, navigating dense tabular data, and performing basic quantitative reasoning. Human solvers average 50 minutes per question, with the majority of time spent locating relevant information across tables and figures within the corpus.

\subsubsection{Setup}

All experiments use Claude Code with Opus 4.5 as the underlying model. Following the data setup described in Section~\ref{methodology:data_setup},

Following the data setup described in Section~\ref{methodology:data_setup}, we partition the benchmark into three disjoint splits: a \textbf{training set}
used for failure detection during evolution, a \textbf{validation set} of 17 examples ($\approx 7\%$) used for frontier selection, and a \textbf{held-out test set} comprising the remaining questions, which is never exposed during evolution. We evaluate three training set sizes: 5\% (12 examples), 10\% (24 examples), and 15\% (36 examples); each evolved for 1.5 epochs. We additionally evaluate a \textbf{skill-merge} configuration, which combines unique skills discovered across independent runs into a single skill library; when skills overlap (identified by matching names or descriptions), we retain the version
from the highest-performing run. All reported accuracies are computed on the held-out test partition unless otherwise noted. We use the fuzzy scoring function provided by OfficeQA, which computes a weighted match across five tolerance levels favoring exact matches. Full scoring details are provided in Appendix~\ref{appendix:data_setup}.

\subsubsection{Results}

\IfFileExists{figures/officeqa_results.pdf}{}{%
  \PackageWarningNoLine{paper}{Missing file figures/officeqa_results.pdf}%
}

\begin{figure}[t]
  \centering
  \includegraphics[width=\linewidth]{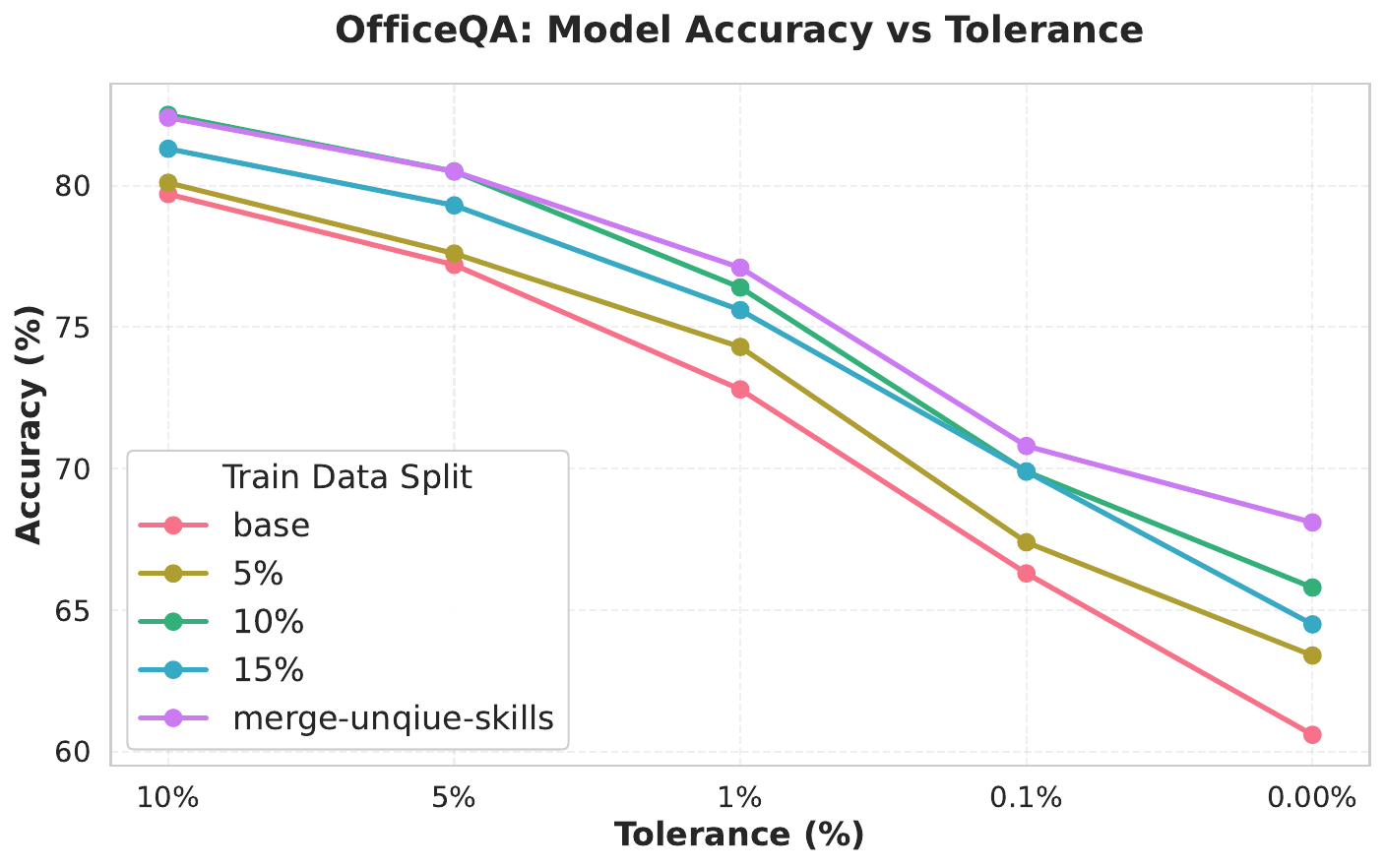}
  \caption{EvoSkill performance OfficeQA benchmark across training splits and tolerance levels. The skill-merge configuration, which combines unique skills from independent runs, achieves the highest exact-match accuracy (67.9\%), a 7.3 percentage point improvement over the baseline (60.6\%).}
  \label{fig:officeqa-results}
\end{figure}

\IfFileExists{tables/officeqa_results.tex}{}{%
  \PackageWarningNoLine{paper}{Missing file tables/officeqa_results.tex}%
}
\begin{table}[ht]
\centering
\caption{Accuracy across tolerance thresholds for different training data splits.}
\label{table:train_split_tolerance}
\vspace{5pt}
\renewcommand{\arraystretch}{1.25}
\setlength{\tabcolsep}{8pt}
\begin{tabular}{lccccc}

\rowcolor{algheader}
\textcolor{algheadertext}{\textsf{\textbf{Train Split/Tol.}}} &
\textcolor{algheadertext}{\textsf{\textbf{0.00\%}}} &
\textcolor{algheadertext}{\textsf{\textbf{0.10\%}}} &
\textcolor{algheadertext}{\textsf{\textbf{1.00\%}}} &
\textcolor{algheadertext}{\textsf{\textbf{5.00\%}}} &
\textcolor{algheadertext}{\textsf{\textbf{10.00\%}}} \\

\arrayrulecolor{algborder}\midrule

                          base & 60.6 & 66.3 & 72.8 & 77.2 & 79.7 \\
\rowcolor{rowalt}          5\% & 63.4 & 67.4 & 74.3 & 77.6 & 80.1 \\
                          10\% & 65.8 & 69.9 & 76.4 & 80.5 & \best{82.5} \\
\rowcolor{rowalt}         15\% & 64.5 & 69.9 & 75.6 & 79.3 & 81.3 \\
\textcolor{accent}{\textsf{merge-unique}} & \best{68.1} & \best{70.8} & \best{77.1} & \best{80.5} & 82.4 \\

\arrayrulecolor{algborder}\bottomrule
\end{tabular}

\vspace{4pt}
{\footnotesize\color{footertext}
\textsf{Tolerance} = allowable relative error.
\colorbox{bestcell}{\textcolor{fn}{\textbf{Blue}}} = best in column.}
\end{table}

Table~\ref{table:train_split_tolerance} presents results across all configurations and tolerance levels. EvoSkill yields consistent improvements over the baseline\footnote{The baseline was independently run \& cross-referenced with authors' most recent result (see: \url{https://github.com/databricks/officeqa/issues/10\#issuecomment-3719842269}).} across all settings. On exact match (0\% tolerance), training on 5\% of the data improves accuracy from 60.6\% to 63.4\% (+2.8\%), while 10\% training data yields 65.8\% (+5.2\%). Performance plateaus beyond 10\%: the 15\% split achieves 64.5\%, slightly below the 10\% run, suggesting diminishing returns or mild overfitting as training data grows.

The skill-merge configuration achieves the strongest result at \textbf{67.9\%} exact match (+7.3\% over baseline), outperforming every individual run. This indicates that skills discovered from independent runs are complementary: each run surfaces different failure modes and corresponding capabilities, and combining them produces a more complete skill library. The improvement pattern is consistent across tolerance levels, with gains ranging from 2.7-4.5\% at stricter tolerances.\footnote{We note that each configuration was evaluated in a single run due to the computational cost of running Opus 4.5 in the evolution loop. Variance analysis across multiple seeds is left to future work.}

\subsubsection{Qualitative Analysis of Discovered Skills}

To illustrate the nature of skills that EvoSkill discovers, we highlight two representative examples from the evolved skill library:

\paragraph{Data Extraction Verification.} EvoSkill discovered a skill enforcing a rigorous protocol for extracting numerical data from Treasury Bulletin tables. The skill is triggered whenever the agent extracts any value from parsed tables, and addresses concrete failure modes identified during evolution: adjacent cell misreads, wrong metric selection, and incorrect time granularity. This skill emerged directly from the Proposer's analysis of cases where the agent retrieved values from neighboring cells or confused similar-sounding metrics.

\paragraph{Quantitative Analysis Methodology.} A second skill provides structured methodology guidance for quantitative financial analysis, including risk calculations, forecasting, currency conversion, and statistical inference. The skill enforces mandatory validation checkpoints before computation, preventing systematic errors from wrong data transformations, date misalignment, or confusion between sample and population statistics.

These examples illustrate that EvoSkill discovers interpretable, domain-relevant skills that target specific failure modes rather than producing opaque optimizations. Additional examples of discovered skills are provided in Appendix~\ref{appendix:skills}.

\subsection{SealQA}

\subsubsection{Benchmark}

SealQA \cite{pham2025sealqaraisingbarreasoning} is a challenge benchmark for evaluating search-augmented language models on fact-seeking questions where web search yields conflicting, noisy, or unhelpful results. Unlike OfficeQA, which tests grounded reasoning over a fixed document corpus, SealQA evaluates an agent's ability to navigate the open web under adversarial retrieval conditions. This makes it a complementary testbed for EvoSkill: the skills required are fundamentally different, centering on search strategy and source verification rather than document parsing and numerical extraction.

\subsubsection{Setup}

  We run EvoSkill on the seal-0 split of SealQA (111 questions) using Claude Code with Opus~4.5 and a 10\% training split, following the partition methodology
  described in Section~\ref{methodology:data_setup}. The remaining questions not used for training or frontier selection form the held-out test set.
  Evolution is run for 1.5 epochs.

\subsubsection{Results}

EvoSkill improves accuracy on SealQA from 26.6\% to \textbf{38.7\%}, an absolute gain of \textbf{12.1\%}. Among the skills discovered, EvoSkill produced a \textit{search-persistence-protocol} that enforces exhaustive search strategies before the agent commits to an answer. The skill requires term interpretation expansion, multi-source verification, and completeness checks, directly addressing the benchmark's core challenge of premature search termination when initial results are noisy or misleading. This result demonstrates that EvoSkill generalizes beyond document-grounded reasoning to search-intensive tasks, discovering qualitatively different skills suited to the target domain.

\subsection{Zero-Shot Skill Transfer}

A key design goal of EvoSkill is that evolved skills, being structured and interpretable, should transfer across tasks without modification. We test this by taking the \textit{search-persistence-protocol} skill evolved on SealQA and applying it zero-shot (with no edits) to BrowseComp \cite{wei2025browsecompsimplechallengingbenchmark}: a benchmark for evaluating browsing agents on challenging fact-seeking questions with short, uniquely correct answers. We evaluate on a stratified sample of 128 examples.

Despite being evolved on a different benchmark with different questions and difficulty characteristics, the transferred skill improves accuracy from 43.5\% to \textbf{48.8\%}, an absolute gain of \textbf{5.3\%}. This provides direct evidence that skills discovered by EvoSkill are not overfit to their training task: the search-persistence-protocol captures a general capability---exhaustive search before committing to an answer---that is broadly useful for fact-seeking tasks regardless of the specific benchmark. This result supports the hypothesis that optimizing at the skill level, rather than at the level of prompts or code, yields more transferable improvements.

\section{Related Work}

\subsection{Agent Skills}

The idea of augmenting agents with reusable, modular capabilities has roots in both embodied AI and software engineering. Voyager \cite{wang2023voyageropenendedembodiedagent} introduced an ever-growing skill library of executable code for an LLM-powered Minecraft agent, where skills are stored as programs in a vector database and retrieved by semantic similarity. Skills in Voyager are discovered through an automatic curriculum and iterative prompting with environment feedback, enabling lifelong learning without parameter updates. More recently, the Agent Skills specification \cite{agentskills} has formalized skills as a portable, open format: each skill is a filesystem directory containing a \texttt{SKILL.md} file with metadata and procedural instructions, optionally bundled with helper scripts and reference materials. This format has been adopted across multiple agent harnesses including Claude Code, the Claude API, and third-party tools \cite{anthropic_skills}\cite{alzubi2026romarecursiveopenmetaagent}. Skills in this paradigm leverage progressive disclosure: metadata is loaded at startup, instructions are read on demand, and scripts are executed without entering the context window enabling agents to maintain many skills with minimal context overhead. Despite the maturity of skill infrastructure, skills today are predominantly hand-authored. EvoSkill addresses this gap by automatically discovering and refining skills through iterative failure analysis, producing artifacts that conform to the same structured skill format.

\subsection{Textual Feedback and Evolutionary Optimization}

A growing body of work explores using natural-language feedback, rather than scalar rewards, to guide iterative improvement of LLM-generated artifacts. Self-Refine \cite{madaan2023selfrefineiterativerefinementselffeedback} demonstrated that a single LLM can improve its own outputs through a generate-critique-refine loop, achieving consistent gains across diverse tasks without additional training. However, Self-Refine operates on individual outputs and does not accumulate knowledge across iterations.

Feedback Descent \cite{lee2025feedbackdescentopenendedtext} formalizes this intuition into a general optimization framework, showing that rich textual feedback from evaluators can drive sustained improvement across domains including molecular design, SVG optimization, and prompt engineering. Feedback Descent maintains a frontier of top-performing candidates and accumulates feedback history, enabling an editor LLM to make increasingly informed revisions. EvoSkill builds directly on this paradigm, applying textual feedback descent to the problem of skill discovery rather than artifact optimization.

On the evolutionary side, AlphaEvolve \cite{novikov2025alphaevolvecodingagentscientific} uses an ensemble of LLMs to evolve entire codebases through an evolutionary loop grounded by automatic evaluation, achieving breakthroughs in algorithm discovery and infrastructure optimization at Google. GEPA \cite{agrawal2026gepareflectivepromptevolution} takes a similar evolutionary approach to prompt optimization within the DSPy framework, using reflective mutation and Pareto-based candidate selection to evolve textual components of complex systems. Both approaches demonstrate the power of evolutionary search with LLM-driven mutations, but they optimize low-level artifacts (code or prompts) that are tightly coupled to specific tasks and models.

EvoSkill differs from these approaches in its level of abstraction. Rather than evolving code or prompts directly, EvoSkill evolves \textit{skills}: structured, reusable capability modules that persist across tasks. This distinction has practical consequences: evolved skills are interpretable, composable, and as we demonstrate transferable to new tasks without modification.

\subsection{Transfer Learning in LLM Agents}

Transfer learning has been extensively studied in the context of neural network fine-tuning, but its application to LLM-based agents remains nascent. In embodied settings, Voyager \cite{wang2023voyageropenendedembodiedagent} showed that a skill library learned in one Minecraft world could be applied to solve novel tasks in a new world, providing early evidence that code-based skills can transfer across environments. More broadly, work on prompt transfer has explored whether optimized prompts generalize across tasks or models, with mixed results---prompts optimized for one setting often degrade when the model or task distribution shifts.

EvoSkill offers a different angle on transfer: because skills are structured as self-contained folders with explicit trigger conditions and procedural instructions, they are decoupled from both the training task and the underlying model. Our experiments provide direct evidence of this: a search-persistence skill evolved on SealQA transfers zero-shot to BrowseComp, yielding a 5.3 percentage point improvement without any modification. This suggests that skill-level optimization may offer a more natural unit of transfer than prompt-level or code-level optimization, though broader investigation across more diverse task pairs is needed.

\section{Conclusion}

We presented EvoSkill, a self-evolving framework that automatically discovers and refines reusable agent skills through iterative failure analysis. By operating at the skill level rather than optimizing low-level artifacts such as prompts or codebases, EvoSkill produces structured, interpretable capabilities that accumulate over iterations and transfer across tasks. Our experiments demonstrate consistent improvements on two distinct benchmarks: OfficeQA (+7.3\%) and SealQA (+12.1\%) using only small training subsets, and we provide direct evidence of zero-shot skill transfer from SealQA to BrowseComp (+5.3\%).

Several directions remain for future work. First, we aim to evaluate EvoSkill across a broader range of domains to better understand the generality of evolved skills and to characterize which skills emerge as \textit{domain-general} (e.g., search persistence, verification protocols) versus \textit{domain-specific} (e.g., treasury table extraction). Second, extending EvoSkill to multi-modal tasks where skills may need to coordinate across vision, code, and language presents a natural next step as coding agents increasingly operate over heterogeneous inputs. Third, the modular structure of evolved skills opens the possibility of building shared skill libraries, where skills discovered on one task can be browsed, composed, and reused by other agents and users. Finally, deeper investigation into the transferability of skills across tasks, models, and agent harnesses, will be critical for realizing the full potential of skill-level optimization as a paradigm for improving coding agents.

\newpage
\bibliographystyle{plainnat}
\bibliography{sample}

\appendix
\section{EvoSkill Generated Skills}
\label{appendix:skills}

The following section shows some of the skills generated using EvoSkill on different agentic tasks.

\subsection{OfficeQA Skills}

The following economic-timeseries-skill is a complex skill that contains both a SKILL.md file and relevant Python scripts to be called with it.

\mdrender{skills/economic-timeseries-analysis/SKILL.md}{skills/economic-timeseries-analysis.md}

\vspace{1em}

\pyrender{skills/economic-timeseries-analysis/scripts/analysis.py}{skills/analysis.py}

\subsection{SealQA}

The following \textit{search-persistence-protocol} skill contains comprehensive instructions for web-search for conclusive answers where conflicting sources may exist. SealQA is considered a challenging benchmark for many multi-agent systems and 

\mdrender{skills/search-persistence-protocol/SKILL.md}{skills/search-persistence.md}

\section{Agent Prompts}
\label{appendix:agent_prompts}

This appendix provides the prompts used for each agent role (placeholders shown).

\subsection{Proposer}
\label{appendix:agent_prompts:proposer}

\mdrender{Proposer Agent Prompt}{appendix/agent-prompts/proposer_placeholder.md}

\subsection{Skill-Builder}
\label{appendix:agent_prompts:skill_builder}

\mdrender{Skill Builder Agent Prompt}{appendix/agent-prompts/skill_builder_placeholder.md}

\subsection{Auto-Grader}
\label{appendix:agent_prompts:auto_grader}
We use the default LLM-as-a-judge \cite{vu2024foundationalautoraterstaminglarge} template provided in the original SealQA task for all auto-grading tasks.

\mdrender{LLM-Judge Prompt}{appendix/agent-prompts/auto_grader_placeholder.md}

\section{Scoring \& Data setup}
\label{appendix:data_setup}
We evaluate agent responses using a deterministic fuzzy matching scorer that returns a binary correctness signal (1.0 or 0.0).
  Given a ground-truth answer $g$ and a predicted answer $p$, the scorer first attempts to extract numerical values from both strings along with
  surrounding textual context (a 20-character window). When both $g$ and $p$ contain numeric content, the scorer normalizes each extracted number
  by detecting unit keywords in the local context (e.g., "million," "billion," "trillion") and compares base values using a relative tolerance
  $\tau$: a prediction is accepted if $|g_{\text{base}} - p_{\text{base}}| / |g_{\text{base}}| \leq \tau$. For the primary evaluation we set $\tau
  = 0$, requiring exact numerical agreement. To avoid spurious matches against incidental year references (e.g., "reported in 2023"), the scorer
  filters candidate numbers in the 1900--2100 range from predictions unless the ground truth itself is a year or contains significant non-numeric
  text. For hybrid answers containing both text and numbers (e.g., "March 1977"), the scorer additionally verifies that key textual elements
  present in the ground truth appear in the prediction via case-insensitive substring matching after stripping unit words and parenthetical
  abbreviations. Multi-number ground-truth answers (i.e., lists) require that all constituent values are recovered in the prediction. For purely
  textual answers, matching is performed via case-insensitive substring containment after normalization (whitespace trimming, quote removal, and
  parenthetical stripping). During the self-improvement training loop, we additionally employ a multi-tolerance scoring variant that computes a
  weighted average over five tolerance levels $\tau \in {0.0, 0.01, 0.025, 0.05, 0.10}$, with weights $w(\tau) = 1/(1 + 20\tau)$ that favor
  stricter thresholds; a weighted score below 0.8 flags the example as a failure for targeted skill refinement.

\section{Environment Branches}
\label{appendix:env_setup}

We manage agent configurations---which we term programs---using a git-backed version control scheme that naturally
  encodes parent--child lineage and supports efficient frontier-based selection. Each program is stored on a dedicated git branch (prefixed
  program/) with a YAML configuration file (.claude/program.yaml) that records the program's name, a pointer to its parent branch, a generation
  counter (i.e., mutation depth from the root), the agent's system prompt, allowed tools, and evaluation metadata including scores. At
  initialization, a base program is created at generation 0 with no parent. At each iteration of the self-improvement loop, the system selects the
  highest-scoring program from a maintained frontier---a bounded set of top-performing programs tracked via git tags (prefixed frontier/)---and
  designates it as the parent. The parent's configuration is then mutated by invoking a proposer agent that analyzes sampled failures, producing a
  child program that inherits all parent attributes (system prompt, tool permissions) while introducing a targeted modification: either a new or edited skill file (in skill-only mode) or a rewritten system prompt (in prompt-only mode). The child is instantiated by checking out the parent
  branch, creating a new branch named iter-{mode}-{n} (where $n$ is the global iteration index), writing the mutated configuration, and committing
  all changes. After the child is evaluated on a held-out validation set, it is admitted to the frontier if either the frontier has not reached its
   maximum capacity $K$ (default $K{=}3$) or its score exceeds that of the current worst frontier member, in which case the weakest member is
  evicted. Children that fail to enter the frontier are discarded---their branches are deleted to prevent repository bloat. This design yields a
  tree-structured search over program space in which each node is a fully reproducible snapshot: lineage can be reconstructed by following parent
  pointers from any program back to the root, and any historical configuration can be restored via a single git checkout. An early-stopping
  criterion halts the loop after a configurable number of consecutive iterations without frontier improvement.

\end{document}